\documentclass{article}

\PassOptionsToPackage{numbers, compress}{natbib}
\usepackage[final]{neurips_2019}

\usepackage{graphicx}  
\usepackage{nameref}
\usepackage{subcaption}
\usepackage{hyperref}       
\usepackage{amsfonts}       
\usepackage{nicefrac}       
\usepackage{microtype}      
\usepackage{amsmath,amssymb}
\usepackage{algorithm}
\usepackage[noend]{algpseudocode}

\newcommand{\argmax}{\arg\!\max}

\usepackage[utf8]{inputenc} 
\usepackage[T1]{fontenc}    
\usepackage{hyperref}       
\usepackage{url}            
\usepackage{booktabs}       
\usepackage{amsfonts}       
\usepackage{nicefrac}       
\usepackage{microtype}      

\title{Tangent Space Separability in Feedforward Neural Networks}

\author{B\'alint Dar\'oczy\\
Institute for Computer Science and Control (SZTAKI)\\
H-1111, Budapest, Hungary\\
\texttt{daroczyb@ilab.sztaki.hu} \\
\And
Rita Aleksziev\\
Institute for Computer Science and Control (SZTAKI)\\
H-1111, Budapest, Hungary\\
\texttt{alexievr@ilab.sztaki.hu} \\
\And
Andr\'as Bencz\'ur\\
Institute for Computer Science and Control (SZTAKI)\\
H-1111, Budapest, Hungary\\
\texttt{benczur@ilab.sztaki.hu} \\
}

\begin{document}

\maketitle

\begin{abstract}
Hierarchical neural networks are exponentially more efficient than their corresponding ``shallow'' counterpart with the same expressive power, but involve huge number of parameters and require tedious amounts of training.
By approximating the tangent subspace, we suggest a sparse representation that enables switching to shallow networks, GradNet after a very early training stage. Our experiments show that the proposed approximation of the metric improves and sometimes even surpasses the achievable performance of the original network significantly even after a few epochs of training the original feedforward network.
\end{abstract}


\section{Introduction}

Recent empirical results of deep hierarchical models go beyond traditional bounds in generalization theory \cite{sontag1998vc,bartlett2003vapnik}, algorithmic complexity \cite{liu2017algorithmic} and local generalization measures \cite{hochreiter1997flat,neyshabur2017exploring,novak2018sensitivity,keskar2016large,kawaguchi2017generalization}. Even simple changes in the models or optimization eventuate significant increase or decrease in performance. Beside exciting empirical phenomenon (e.g. larger networks generalize better, different optimization with zero training error may generalize differently \cite{neyshabur2018towards}) and theoretical developments (e.g. flatness of the minimum can be changed arbitrarily under some meaningful conditions via exploiting symmetries \cite{dinh2017sharp}) our understanding of deep neural networks still remain incomplete \cite{neyshabur2018towards}. There are several promising ideas and approaches inspired by statistical physics \cite{rolnick2017power}, tensor networks \cite{stoudenmire2016supervised} or in our case particularly by differential geometry \cite{amari1996neural,ollivier2015riemannian,kanwal2017comparing}. 

We investigate the connection between the structure of a neural network and Riemannian manifolds to utilize more of their potential. In a way, many of the existing machine learning problems can be investigated as statistical learning problems. Although information geometry \cite{amari1996neural} plays an important role in statistical learning, the geometrical properties of target functions both widely used and recently discovered, along with those of the models themselves, are not well studied. 

Over the parameter space and the error function we can often determine a smooth manifold \cite{ollivier2015riemannian}. In this paper we investigate the tangent bundle of this manifold in order to take advantage of specific Riemannian metrics having unique invariance properties \cite{Cencov1982,campbell1986extended}. We use the partial derivatives in the tangent space as representation of data points. The inner products in the tangent space are quadratic, therefore if we separate the samples with a second order polynomial, then the actual metric will be irrelevant. 

The geometrical property of the underlying manifold was used for optimizing generative models \cite{rifai2011manifold} and as a general framework for optimization in \cite{ollivier2015riemannian,zhang2016riemannian}, neither of them utilize the tangent space as representation. The closest to our method are \cite{jaakkola1999exploiting} where the authors used the diagonal of the Fisher information matrix of some generative probability density functions in a kernel function for classification. Martens and Grosse \cite{martens2015optimizing} approximated the Amari's natural gradient \cite{amari1996neural} with block-partinioning the Fisher information matrix. The authors show that, under some assumptions the proposed Kronecker-factored Approximate Curvature (K-FAC) method still preserves some invariance results related to Fisher information. Closed formula for Gaussian Mixtures was proposed in \cite{perronnin2007fisher}. Our contributions are the following:

\begin{itemize}
\item We suggest an approximation algorithm for the inner product space, in case of weekly trained networks of various sparsities.
\item We give heuristics for classification tasks where the coefficients of the monomials in the inner product are not necessarily determined.
\end{itemize}

Our experiments were done on the CIFAR-10 \cite{krizhevsky2009learning} and MNIST \cite{lecun-mnisthandwrittendigit-2010} data sets. We show that if we determine the metric of the underlying feedforward neural network in an early stage of learning (after only a few epochs), we can outperform the fully trained network by passing the linearized inner product space to a shallow network.

\section{Tangent space and neural networks}

Feed-forward networks with parameters $\theta$ solve the optimization problem
$\min_\theta f(\theta) = \mathbb{E}_{\mathcal{X}}[l(x;\theta)]$ where $l(x;\theta)$ is usually a non-convex function of configuration parameters $\theta$. In case of discriminative models the loss function depends on the target variable as well: $l(x;c,\theta)$. 

First, we define a differential manifold $(\mathcal{M})$ 
based on $l(x;\theta)$ by assigning a tangent subspace and a particular vector to each configuration point, a parameter sample pair $(x,\theta)$. In case of Riemann the metric induced via an inner product $g_{x,\theta}: T_{x,\theta}M \times T_{x,\theta}M \rightarrow \mathbb{R}$ where $\{x,\theta\} \in \{\mathcal{X},\Theta\} \subset \mathcal{M}$. If we minimize over a finite set of known examples, then the problem is closely related to the empirical risk minimization and loglikelihood maximization. 


The parameter space of continuously differentiable feedforward neural networks (CDFNN) usually has a Riemannian metric structure \cite{ollivier2015riemannian}. Formally, let $X=\{x_1,..,x_T\}$ be a finite set of known observations with or without a set of target variables $Y=\{y_1,..,y_T\}$ and a directed graph $N=\{V,E\}$ where $V$ is the set of nodes with their activation functions and $E$ is the set of weighted, directed edges between the nodes. Let the loss function $l$ be additive over $X$. Now, in case of generative models, the optimization problem has the form $\min_{f \in \mathcal{F}_N} l(X;f) = \min_{f \in \mathcal{F}_N} \frac{1}{T}\sum_i^T l(x_i;f)$ where $\mathcal{F}_{N}$ is the class of neural networks with structure $N$. Optimization can be interpreted as a ``random walk'' on the manifold with finite steps defined by some transition function between the points and their tangent subspaces. 

The general constraint about Riemannian metrics is that the metric tensor should be symmetric, positive definite and the inner product in the tangent space assigned to any point on a finite dimensional manifold has to be able to be formalized as $<x,x>_{\theta} = dx^T G_{\theta} dx = \sum_{i,j} g_{i,j}^{\theta} dx^i dx^j$. The metric $g_{\theta}$ varies smoothly by $\theta$ on the manifold and is arbitrary given the conditions. 

\subsection{GradNet}

Let our loss function $l(x;\theta)$ be a smooth, positive, parametric real function where $x \in \mathbb{R}^d$ and $\theta \in \mathbb{R}^n$. We define a class of $n \times n$ positive semi-definite matrices as 

\begin{equation}
\label{eq:LC_dot}
h_{\theta}(x) = \nabla_{\theta} l(x;\theta) \otimes \nabla_{\theta} l(x;\theta)
\end{equation}

where $ \nabla_{\theta} l(x;\theta) = \{\frac{\partial l(x;\theta)}{\partial \theta_1},..,\frac{\partial l(x;\theta)}{\partial \theta_n}\}$. 
%
%
%
Using eq.~(\ref{eq:LC_dot}) we can determine a class of Riemannian metrics 
\begin{equation}
G = g_{X}(h_{\theta}(x))
\label{eq:riemann-metric}
\end{equation}
where $g_X$ is a quasi arithmetic mean over $X$. For example, if $g_X$ is the arithmetic mean, then the metric is $ G_{\theta}=\mathbf{AM}_{X} [\nabla _\theta l(x_i|\theta)\nabla _\theta l(x_i|\theta)^T]$ and we can approximate it with a finite sum as $G_{\theta}^{kl} \approx \sum_{i} \omega_i
\left(\frac{\partial}{\partial \theta_k} l(x_i|\theta)\right)
\left(\frac{\partial}{\partial \theta_l} l(x_i|\theta)\right)$ with some importance $\omega_i$ assigned to each sample. Through $G$, the tangent bundle of the Riemannian manifold induces a normalized inner product (kernel) at any configuration of the parameters formalized for two samples $x_i$ and $x_j$ as 

\begin{equation}
\label{eq:K_base}
<x_i,x_j>_{\theta} = \nabla_{\theta} l(x_i;\theta)^T G_{\theta}^{-1} \nabla_{\theta} l(x_j;\theta)
\end{equation}

where the inverse of $G_{\theta}$ is positive semi-definite since $G_{\theta}$ is positive semi-definite.  

\begin{figure}
\centerline{\includegraphics[scale=.2]{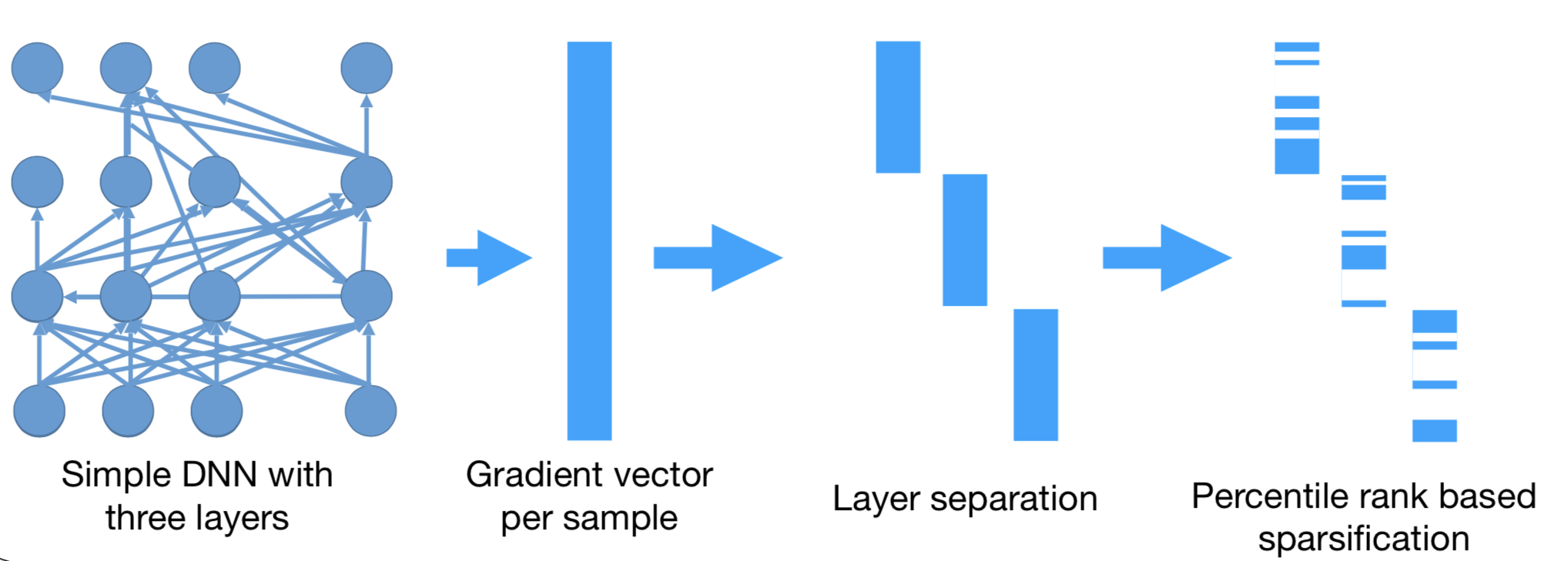}}
\caption{Layer-by-layer per sample sparsification by percentile ranking}
\label{fig:sparse}
\end{figure}

The quadratic nature of the Riemannian metrics is a serious concern due to high dimensionality of the tangent space. There are several ways to determine a linear inner product: decomposition or diagonal approximation of the metric, or quadratic flattening. Due to high dimensionality, both decomposition and flattening can be highly inefficient, although flattening can be highly sparse in case of sparsified gradients. Our main idea is to combine per sample sparsification with quasi-blockdiagonal approximation, shown in Fig.~\ref{fig:sparse}. 

\begin{figure}
\centering
\includegraphics[width = .4\textwidth]{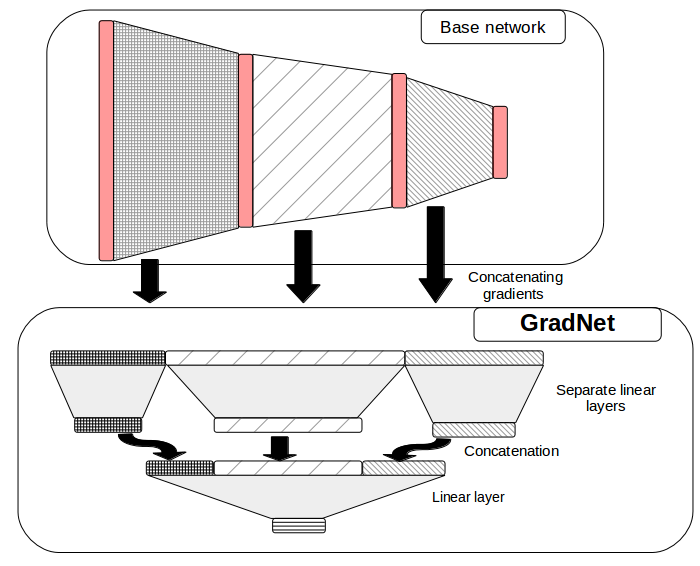}
\caption{GradNet}
\label{fig:gradnet}
\end{figure}

But before we trade on our new sparse representation, we have to handle another problem. Since our models are discriminative and not generative, the loss surfaces are not known in absence of the labels.  Hence we define GradNet, Fig.~\ref{fig:gradnet}, a multi-layer network  over the tangent space, as  $h_{\mbox{\scriptsize GradNet}}(x;l(x,\hat{c};\theta))$ with the assumption  that the final output of the network after training is $\argmax_{c} \sum_{\hat{c}} h_{\mbox{\scriptsize GradNet}}(x;l(x;\hat{c},\theta)$. 

Results in \cite{denil2013predicting,denton2014exploiting,choromanska2015loss} indicate high over-parametrization and redundancy in the parameter space, especially in deeper feedforward networks, therefore the outer product structure is highly blocked particularly in case of ReLU networks and sparsified gradients. 

\begin{figure}
\centerline{\includegraphics[scale=.2]{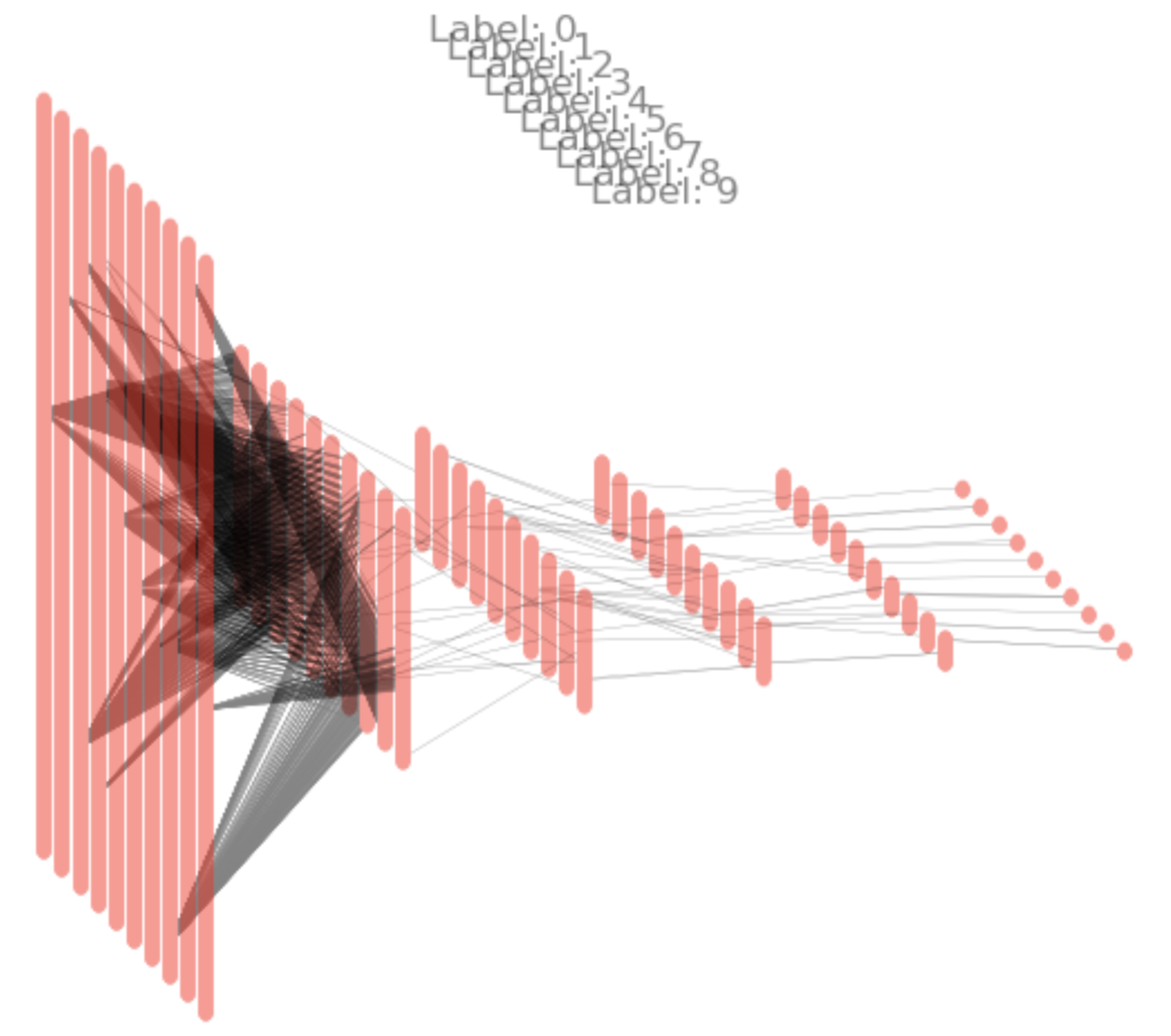}}
\caption{Important edges in the gradient graph of the MNIST network.}
\label{fig:grad1}
\end{figure}

Let us consider a multi-layer perceptron with rectified linear units and a gradient graph with sparsity factor $\alpha$ corresponding to the proportion of the most important edges in the gradient graph derived by the Hessian for a particular sample. The nodes are corresponding to the parameters and we connect two nodes with a weighted edge if their value in the Hessian matrix is nonzero and their layers are neighbors. We sparsify the gradient graph layer-by-layer by the absolute weight of the edges. As shown in Fig.~\ref{fig:grad1}, the resulting sparse graph describes the structure of the MINST task well: for each label 0--9, a few nodes are selected in the first layer and only a few paths leave the second layer. 

In order to avoid having too much parameters to train, we chose the hidden layer of our network to have a block-like structure demonstrated in Figure \ref{fig:gradnet}. This model is capable of capturing connections between gradients from adjacent layers of the base network.

The final algorithm is 

\begin{algorithm}[H]
\caption{Training procedure of GradNet}
\textbf{Input:} Pre-trained model with parameter $\theta$, dataset $D$, GradNet $N$, normalization method $n$, number of epochs $t$ and sparsification coefficient $\alpha$

\textbf{Output:} Trained GradNet

\begin{algorithmic}[1]
\Procedure{Train}{$M, D, N, n, t$}

\For{\texttt{epoch from 1 to t}}
\For{\texttt{$batch$ in $D$}}
\State $X \gets augmentation(batch)$
\State $c \gets$ real labels for each data point in $batch$
\State $\hat{c} \gets$ random labels for each data point in $batch$ 
\State $X_g \gets \nabla_{\theta}(l(x;\hat{c},\theta))$ for each data point in the $batch$
\State $\hat{X_g} \gets n(X_g,\alpha)$  \Comment{normalization and sparsification}
\State $N \gets update(N, \hat{X_g}, c)$ \Comment{update network with normalized gradients}
\EndFor
\EndFor
\State \textbf{return} $N$\Comment{Return trained $N$}

Prediction for data point $x$: $\argmax_{c} \sum_{\hat{c}} N(n(\nabla_{\theta}l(x;\hat{c},\theta)))$


\EndProcedure
\end{algorithmic}
\end{algorithm}
\section{Experiments}

In our first experiment we set Markov Random Fields (MRF), particularly Restricted Boltzmann Machines (RBM)\cite{hinton2002training} with $16$ and $64$ hidden units on the first half of the MNIST \cite{lecun1998gradient} training set and calculated the normalized gradient vectors based on the Hammersley-Clifford theorem \cite{hammersley1971markov}. We used the RBMs output and the normalized gradient vectors within a linear model. The results (see \nameref{sec:appendix}) show that the normalized gradient vector space performed very similar after some initialization with $1k$ sample and after training while the original latent space performed poorly immediately after initialization. 

We trained several traditional Convolutional Neural Networks (CNN) and descendants such as residual \cite{he2016deep} and dense networks \cite{iandola2014densenet} over the CIFAR-10 dataset \cite{krizhevsky2009learning} as the base model for GradNet. In order to find the optimal architecture, the right normalization process and regularization technique, and the best optimization method, we experimented with a large number of setups (see \nameref{sec:appendix}). Interestingly, the GradNet surprassed the performance of the underlying CNN, at some settings even after only one epoch. Source codes are available at \url{https://github.com/daroczyb/gradnet}. 
\section{Conclusions}

 By approximation of the inner product, we showed promising results with our GradNet network in the sparsified gradient space. GradNet outperformed the original network even if built from a few epochs of the original network. In the future, we would like to extend our method to Hessian metrics and further investigate sparsity and possible transitions to less complex manifolds via pushforward and random orthogonal transformations.

\section{Acknowledgement}
The publication was supported by the Hungarian Government project 2018-1.2.1-NKP-00008: Exploring the Mathematical Foundations of Artificial Intelligence, by the Higher Education Institutional Excellence Program, and by the Momentum Grant of the Hungarian Academy of Sciences. B.D. was supported by an MTA Premium Postdoctoral Grant 2018.

\bibliographystyle{apalike}
\bibliography{tangent,hessian}

 \section*{Appendix}
 \label{sec:appendix}

 We measured the performance of various GradNet models on the gradient space of a CNN trained on the first half of the CIFAR-10 \cite{krizhevsky2009learning} training dataset. We used the other half of the dataset and random labels to generate gradient vectors to be used as training input for the GradNet. In the testing phase we use all of the gradient vectors for every data point in the test set, we give them all to the network as inputs, and we define the prediction as the index of the maximal element in the sum of the outputs.

During our experiments, as a starting point we stopped the underlying original CNN at $0.72$ accuracy and compared the following settings. 
\begin{itemize}
\item Regarding \underline{\smash{regularization}}, we considered using dropout, batch normalization, both of them together, or none.
\item We experimented with SGD and Adam \underline{\smash{optimization methods}}.
\item Since we suspected that not all coordinates of the gradients are equally important, we only used the elements of large absolute value making the process computationally less expensive. We kept the elements of absolute value greater than the \underline{\smash{$q$-th percentile}} of the absolute value vector, and we tested our model setting this $q$ value for 99, 95, 90, 85, 80 and 70. We also tried a method where we pre-computed the indices of the most important 10$\%$ of the values for each label, and used this together with the above technique.
\item In order to determine the exact \underline{structure} of the GradNet, we tried layers and blocks of different sizes. These models differ only in the size and partition of the middle layer, which were the following in our tests: 5+25+10; 20+100+40; 10+50+20; 5+100+25; 10+200+50; 20+400+100. 
\item In terms of \underline{normalization}, we ran tests using standard norm with and without L2-norm following it; scale norm; power norm with exponents $\frac{1}{8}$, $\frac{1}{4}$, $\frac{1}{2}$, and $2$; and scale norm followed by power norm with exponent $\frac{1}{2}$.
\end{itemize}

\begin{figure}
\begin{minipage}[c]{1\linewidth}
\includegraphics[width = \textwidth]{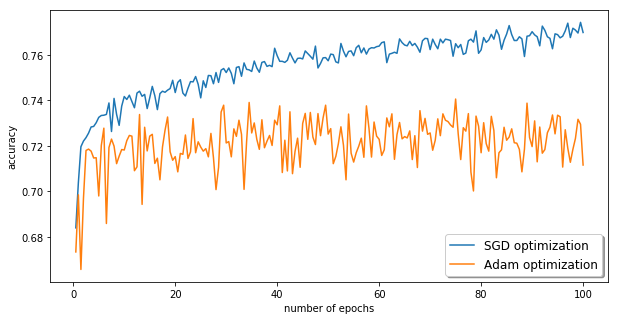}
\caption{Optimization methods}
\label{fig:opt}
\end{minipage}
\hfill
\begin{minipage}[c]{1\linewidth}
\includegraphics[width=\textwidth]{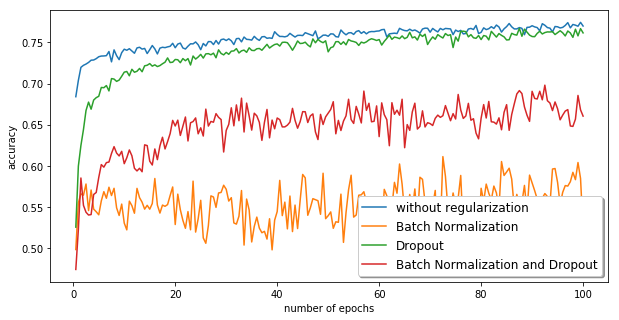}
\caption{Regularization methods}
\label{fig:reg}
\end{minipage}
\end{figure}

\begin{figure}
\begin{minipage}[c]{1\linewidth}
\includegraphics[width = \textwidth]{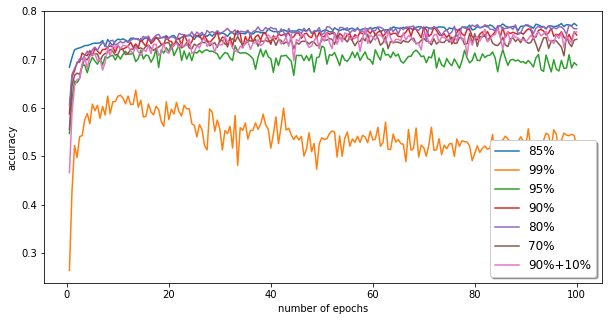}
\caption{Selection percentile}
\label{fig:percent}
\end{minipage}
\hfill
\begin{minipage}[c]{1\linewidth}
\includegraphics[width=\textwidth]{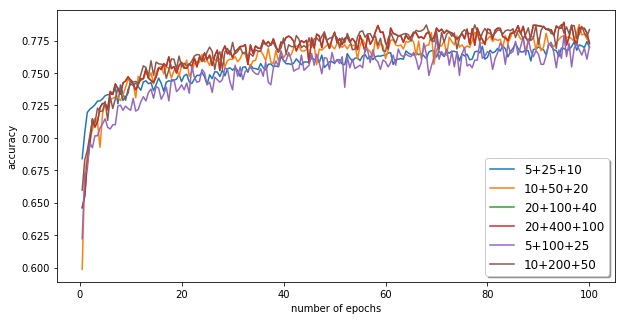}
\caption{Structure}
\label{fig:structure}
\end{minipage}
\end{figure}

\begin{figure}
\begin{minipage}[c]{1\linewidth}
\includegraphics[width = \textwidth]{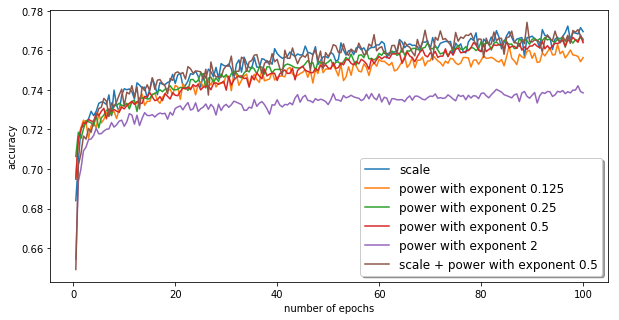}
\caption{Normalization with structure 5+25+10}
\label{fig:norm1}
\end{minipage}
\hfill
\begin{minipage}[c]{1\linewidth}
\includegraphics[width=\textwidth]{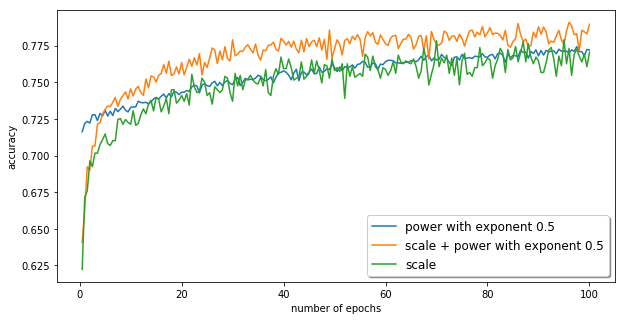}
\caption{Normalization with structure 5+100+25}
\label{fig:norm2}
\end{minipage}
\end{figure}

  Learning curves for the different networks are presented in Figures \ref{fig:opt} - \ref{fig:norm2}. We observed that SGD gives a better performance than Adam (Fig.~\ref{fig:opt}), and that regularization is not needed (Fig.~\ref{fig:reg}). We also found that it is sufficient to use the elements of each gradient vector that are greater than the 85-th percentile of all of the absolute values in the vector (Fig.~\ref{fig:percent}). Regarding structure, the best-performing GradNet was the one with hidden layer of size 130 partitioned into sublayers of sizes 5, 100 and 25 (Fig.~\ref{fig:structure}). Out of all the considered normalization methods, the scale norm and the power norm together gave the most satisfactory outcome (Fig.~\ref{fig:norm1},\ref{fig:norm2}). 

\begin{table}
    \centering
    \label{tab:rbm}
    \caption{Performance measure of the normalized gradient based on RBM.}
\begin{tabular}{| c | c | c |}
    \hline
    \multicolumn{3}{|c|}{MNIST}\\
    \hline
    \#hidden & Base RBM & GradNet \\ \hline
    16  & 0.6834 & 0.9675 \\ \hline
    16 & 0.8997 & 0.9734 \\ \hline
	64 & 0.872 & 0.9822 \\ \hline
	64 & 0.9134 & 0.9876 \\ \hline
    \end{tabular}
\end{table}

To show the performance of the GradNet with these particular settings, we took snapshots of a CNN at progressively increasing levels of pre-training, and we trained the GradNet on the gradient sets of these networks. We ran these tests using a CNN trained on half of the CIFAR dataset and with one trained on half of MNIST. Table \ref{table:eval} shows the accuracies of all the base networks together with the accuracies of the corresponding GradNets. 

\begin{table}
    \centering
    \caption{Performance measure of the improved networks.}
    \begin{tabular}{| c | c | c |}
    \hline
    \multicolumn{3}{|c|}{CIFAR}\\
    \hline
    Base NN & GradNet & Gain \\ \hline
    0.79  & 0.8289 & +4.9\% \\ \hline
    0.76 & 0.8201 & +7.9\% \\ \hline
	0.74 & 0.8066 & +9\% \\ \hline
	0.72 & 0.7936 & +10.2\% \\ \hline
	0.68 & 0.7649 & +12.5\% \\ \hline
	0.65 & 0.7511 & +15.5\% \\ \hline
	0.62 & 0.7274 & +17.3\% \\ \hline
	0.55 & 0.7016 & +27.5\% \\ \hline
	0.51 & 0.6856 & +34.4\% \\ \hline
	0.49 & 0.678 & +38.3\% \\ \hline
    \end{tabular}
    \begin{tabular}{| c | c | c |}
    \hline
    \multicolumn{3}{|c|}{MNIST}\\
    \hline
    Base NN & GradNet & Gain \\ \hline
    0.92  & 0.98 & +6.5\% \\ \hline
    0.96 & 0.9857 & +2.7\% \\ \hline
	0.9894 & 0.9914 & +0.2\% \\ \hline
    \end{tabular}
    
    \vspace{5 pt}
    \label{table:eval}

\end{table}

\end{document}